# HANDWRITTEN 'BANGLA' ALPHABET RECOGNITION USING AN MLP BASED CLASSIFIER

Subhadip Basu [+], Nibaran Das*, Ram Sarkar*, Mahantapas Kundu*,
Mita Nasipuri* [ξ], Dipak Kumar Basu*

[+] Computer Sc. & Engg. Dept., MCKV Institute of Engineering, Liluah, Howrah-711204, India.
*Computer Sc. & Engg. Dept., Jadavpur University, Kolkata-700032, India.

[ξ] Corresponding Author; e-mail: nasipuri@vsnl.com

*Abstract:*
*The work presented here involves the design of a Multi Layer Perceptron (MLP) based classifier for recognition of handwritten Bangla alphabet using a 76 element feature set Bangla is the second most popular script and language in the Indian subcontinent and the fifth most popular language in the world. The feature set developed for representing handwritten characters of Bangla alphabet includes 24 shadow features, 16 centroid features and 36 longest-run features. Recognition performances of the MLP designed to work with this feature set are experimentally observed as 86.46% and 75.05% on the samples of the training and the test sets respectively. The work has useful application in the development of a complete OCR system for handwritten Bangla text.*

# HANDWRITTEN 'BANGLA' ALPHABET RECOGNITION USING AN MLP BASED CLASSIFIER

Subhadip Basu [+], Nibaran Das*, Ram Sarkar*, Mahantapas Kundu*,
Mita Nasipuri* [ξ], Dipak Kumar Basu*

[+] Computer Sc. & Engg. Dept., MCKV Institute of Engineering, Liluah, Howrah-711204, India.
*Computer Sc. & Engg. Dept., Jadavpur University, Kolkata-700032, India.

[ξ] Corresponding Author; e-mail: nasipuri@vsnl.com

*Abstract:*

*The work presented here involves the design of a Multi Layer Perceptron (MLP) based classifier for recognition of handwritten Bangla alphabet using a 76 element feature set Bangla is the second most popular script and language in the Indian subcontinent and the fifth most popular language in the world. The feature set developed for representing handwritten characters of Bangla alphabet includes 24 shadow features, 16 centroid features and 36 longest-run features. Recognition performances of the MLP designed to work with this feature set are experimentally observed as 86.46% and 75.05% on the samples of the training and the test sets respectively. The work has useful application in the development of a complete OCR system for handwritten Bangla text.*

## 1. INTRODUCTION

Optical Character Recognition (OCR) is still an active area of research, especially for handwritten text. Success of the commercially available OCR system is yet to be extended to handwritten text. It is mainly due to the fact that numerous variations in writing styles of individuals make recognition of handwritten characters difficult. Past work on OCR of handwritten alphabet and numerals has been mostly found to concentrate on Roman script [3], related to English and some European languages, and scripts related to Asian languages like Chinese [2], Korean, and Japanese.

Among Indian scripts, Devnagri, Tamil, Oriya and Bangla have started to receive attention for OCR related research in the recent years. Out of these, Bangla is the second most popular script and language in the Indian subcontinent. As a script, it is used for Bangla, Ahamia and Manipuri languages. Bangla, which is also the national language of Bangladesh, is the fifth most popular language in the world. So is the importance of Bangla both as a script and as a language. But evidences of research on OCR of handwritten Bangla characters, as observed in the literature, are a few in numbers.

Two of the important research contributions relating to OCR of Bangla characters involve a multistage approach developed by Rahman et al. [1] and an MLP classifier developed by Bhowmik et al. [4]. The major features used for the *multistage approach* include *Matra*, upper part of the character, disjoint section of the character, vertical line and double vertical line. And, for the MLP classifier, the feature set is constructed from the stroke feature of characters. The data set used for testing recognition performances of the multistage approach was not of considerable size as it included characters of 49 different classes collected from only 20 different writers. Compared to this, a moderately large size data set of 25,000 samples, collected from different sections of population, is used for testing performances of the MLP classifier. The size of the input feature vector chosen for the work is 200.

Not only because of numerous variation of writing styles of different individuals but also for the complex nature of Bangla alphabet, automatic recognition of handwritten Bangla characters still poses some potential problems to the researchers. Compared to Roman alphabet, basic Bangla alphabet consists of a much larger number of characters. The number of characters in basic Bangla alphabet is 50. And some characters therein resemble pair wise so closely that the only sign of small difference left between them is a period or a small line. Handwritten samples of all 50 symbols of basic Bangla alphabet are shown in Fig 1.

In the light of above facts, the present work considers a feature set of 76 features, smaller than the one previously used by Bhowmik et al. (2005), for recognition of handwritten Bangla alphabet using an MLP based classifier. Due to the non-availability of some standardized data set, recognition performances of the present work cannot be compared with the others. However, for the currently available data set, the work shows satisfactory recognition performances with some specific suggestions to be tried in future for further performance enhancement.

(a)

(b)

Fig 1. BMP images of 50 handwritten characters of basic Bangla alphabet.
    (a). Vowels of Bangla script
    (b). Consonants of Bangla script

## 2. THE FEATURE SET

The feature set selected for the present work consists of 76 features, which include 24 shadow features, 16 centroid features and 36 longest-run features. The features are computed from 64x64 pixel size binary images of alphabetic characters.

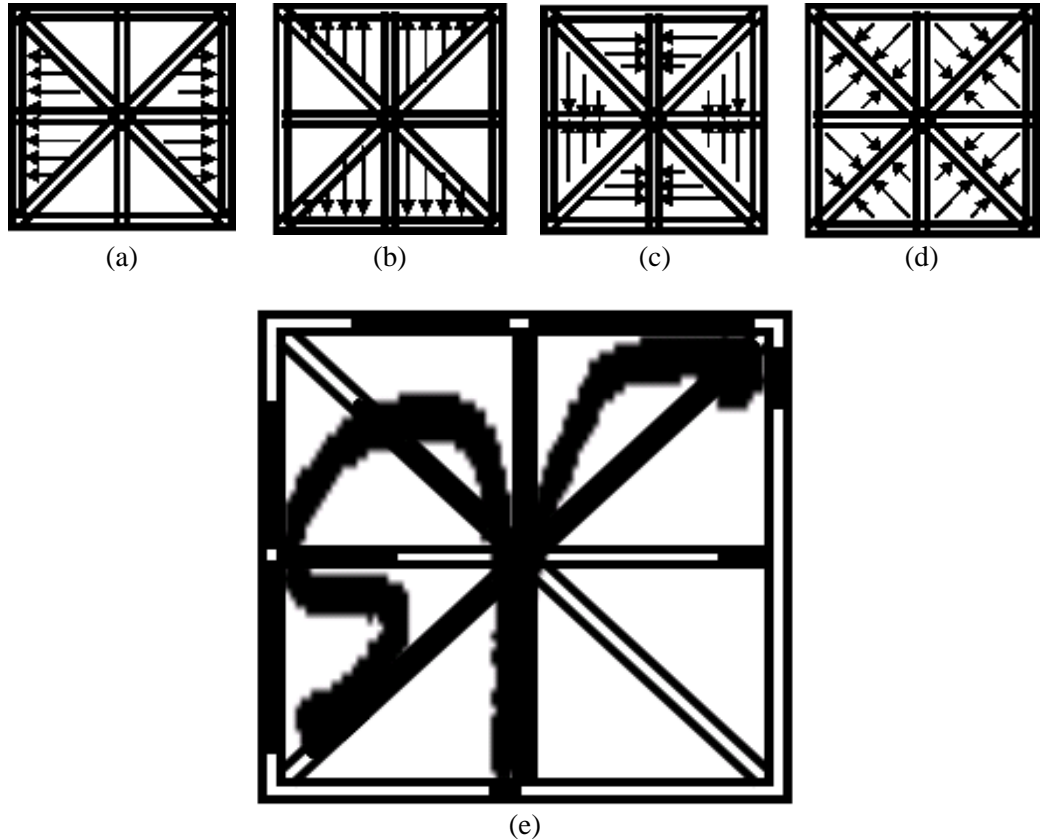

Fig 2. An illustration for shadow features.
(a-d) Direction of fictitious light rays as assume for taking the projection of an image segment on each side of all octants.
(e) Projection of a sample image.

## 2.1. Shadow Features

For computing shadow features, each character image is enclosed within a minimal square, divided into eight octants, as shown in Fig. 2. Lengths of projections of character images on three sides of each octant are then computed. Finally lengths of all such projections on each of the 24 sides of all octants are summed up to produce 24 shadow features of the character image under consideration. For taking the projection of an image segment on one side of an octant, existence of a fictitious light source in the opposite side is assume. Directions of light rays so assume for each side of all octants in a minimal square are shown in Fig 2(a-d). Shadow features of a sample character image are illustrated also in Fig 2(e).

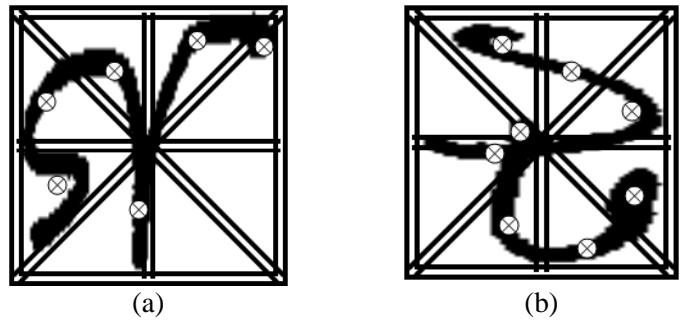

Fig 3(a-b). Centroid features of two different character images

## 2.2. Centriod Features

Coordinates of centroids of black pixels in all the 8 octants of a digit image are considered to add 16 features in all to the feature set. Fig 3 (a-b) show approximate locations of all such centroids on two different character images. It is noteworthy how these features can be of help to distinguish the two images.

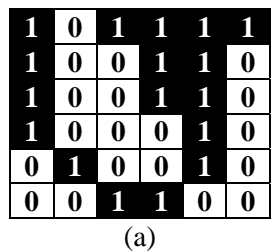

(a)

|  | Length of the Longest Bar |  |  |  |  |  |
|---|---|---|---|---|---|---|
| 1 | 0 | 4 | 4 | 4 | 4 | 4 |
| 1 | 0 | 0 | 2 | 2 | 0 | 2 |
| 1 | 0 | 0 | 2 | 2 | 0 | 2 |
| 1 | 0 | 0 | 0 | 1 | 0 | 1 |
| 0 | 1 | 0 | 0 | 1 | 0 | 1 |
| 0 | 0 | 2 | 2 | 0 | 0 | 2 |

Sum = 12

(b)

Fig 4. An illustration for computation of the row wise longest–run feature.
(a) The portion of a binary image enclosed within a rectangular region.
(b) Every pixel position in each row of the image is marked with the length of the longest bar that fits consecutive black pixels along the same row.

## 2.3. Longest-run Features

For computing longest-run features from a character image, the minimal square enclosing the image is divided into 9 overlapping rectangular regions. Coordinates (r, c) of top left corners of all these regions are given as follows: {(r, c) | r=0, h/4, 2h/4 and c=0, w/4, 2w/4}, where h and w denote the height and the

width of the minimal square respectively. In each such rectangular region, 4 longest-run features are computed row wise, column wise and along two of its major diagonals.

The row wise longest-run feature is computed by considering the *sum* of the *lengths* of the *longest bars* that fit consecutive black pixels along each of all the rows of a rectangular region, as illustrated in Fig. 4. The three other longest-run features are computed in the same way but along the column wise and two major diagonal wise directions within the rectangle separately. Thus, in all, 9x4=36 longest-run features are computed from each character image.

## 3. THE MLP CLASSIFIER

The MLP is a special kind of Artificial Neural Network (ANN). ANNs are developed to replicate *learning* and *generalization* abilities of human's behavior with an attempt to model the functions of *biological neural networks* of the human brain.

Architecturally, an MLP is a feed-forward layered network of *artificial neurons*. Each artificial neuron in the MLP computes a *sigmoid function* of the weighted sum of all its inputs. An MLP consists of one *input layer*, one *output layer* and a number of *hidden* or intermediate *layers*, as shown in Fig 5. The output from every neuron in a layer of the MLP is connected to all inputs of each neuron in the immediate next layer of the same. Neurons in the input layer of the MLP are all basically dummy neurons as they are used simply to pass on the input to the next layer just by computing an identity function each.

The numbers of neurons in the input and the output layers of an MLP are chosen depending on the problem to be solved. The number of neurons in other layers and the number of layers in the MLP are all determined by a trial and error method at the time of its *training*. An ANN requires training to learn an unknown input-output relationship to solve a problem.

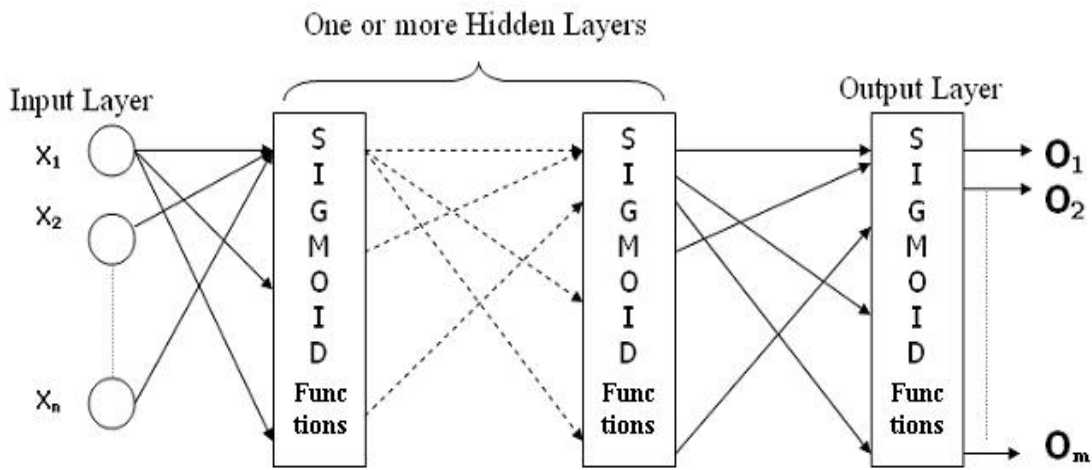

Fig. 5. A block diagram of an MLP shown as a feed forward neural network.

Depending on the models of ANNs, training is performed either under supervision of some teacher (i.e., with labeled data of known input-output responses) or without supervision. The MLP to be used for the present work requires supervised training. During training of an MLP *weights* or strengths of neuron-to-neuron connections, also called *synapses*, are iteratively tuned so that it can respond appropriately to all training data and also to other data, not considered at the time of training. Learning and generalization abilities of an ANN is determined on the basis of how best it can respond under these two respective situations.

The MLP classifier designed for the present work is trained with the Back Propagation (BP) algorithm. It minimizes the *sum of the squared errors* for the training samples by conducting a *gradient descent* search in the *weight space*. The number neurons in a hidden layer in the same are also adjusted during its training.

The problem of *pattern classification* involves two successive transformations as follows:

M→F→D

where, M, F and D stand for the measurement space, the feature space and the decision space respectively. Once a feature set is fixed up, it is left with the design of a mapping (δ) as follows:

δ: F→D

ANNs with their learning and generalization abilities can approximate a general class of functions given below.

$$f : \mathbb{R}^n \to \mathbb{R}$$

Pattern classification with ANNs requires approximating δ as a *discrete valued function* shown below.

$$\delta : \mathbb{R}^n \to \{1, 2, \ldots m\}$$

where, n and m denotes the number of features and the number of pattern classes respectively. So an ANN based pattern classifier requires n number of neurons in the input layer and m number of neurons in the output layer. Conventionally 1-out-of-m representation is used for its output.

Table 1. Recognition performances of the MLP with different numbers of neurons in the hidden layers

| *No of Hidden neurons* | 35 | 40 | 45 | 50 | 55 | 60 | 65 | 70 | 75 |
|---|---|---|---|---|---|---|---|---|---|
| **Percentage recognition rate on the training samples** | 80.35 | 80.93 | 82.8 | 84.7 | 85.70 | 86.46 | 87.36 | 86.69 | 85.65 |
| **Percentage recognition rate on test samples** | 70.35 | 70.70 | 71.6 | 73.15 | 74.2 | 75.05 | 73.65 | 74.7 | 72.00 |

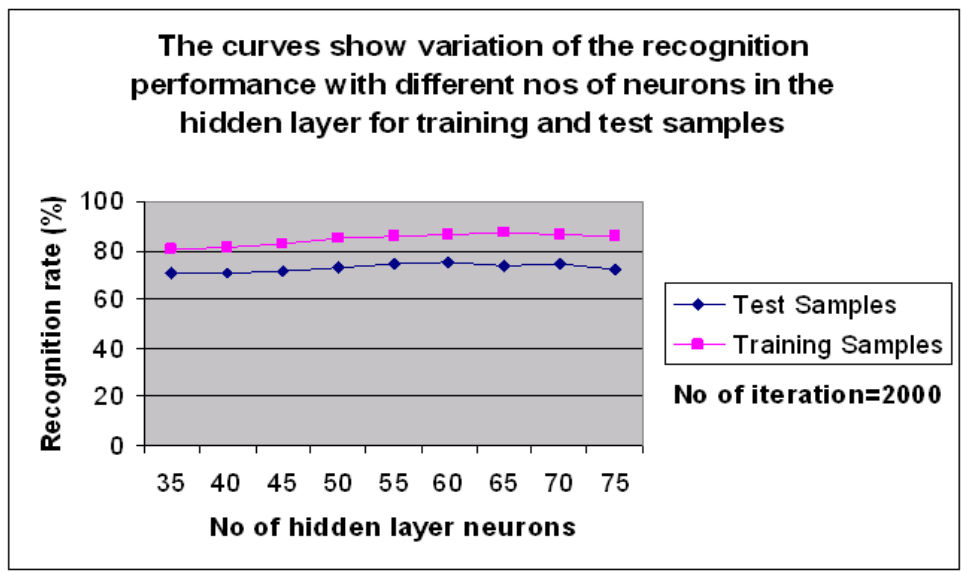

Fig 6. Curves show variation of recognition performances of the MLP as the number of neurons in its hidden layer is increased.

# 4. RESULTS AND DISCUSSION

For preparation of the *training* and the *test sets* of samples, a database of 10,000 alphabetic character samples is formed by collecting optically scanned handwritten characters specimens of 50 alphabetic symbols from each of 200 people of different age groups and sexes. A *training set* of 8000 samples and a *test set* of 2000 samples are then formed through random selection of character samples of each class from the initial database in equal numbers. All these samples are scaled to 64x64 pixel images first and then converted to *binary images* through thresholding.

For the present work, a single layer MLP, i.e., an MLP with one hidden layer is chosen. This is mainly to keep the computational requirement of the same low without affecting its function approximation capability. According to Universal Approximation theorem [5], a single hidden layer is sufficient to compute a uniform approximation to a given training set.

To design an MLP for classification of handwritten alphabetic characters, several runs of BP algorithm with *learning rate* ($\eta$) = 0.8 and *momentum term* ($\alpha$)=0.7 are executed for different numbers of neurons in its hidden layer. Recognition performances of the MLP on the training and the test sets observed from this experimentation are given in Table1.

Curves showing variation of the Recognition performance of the MLP, for both the test and the training sets, with increase in the number of neurons in its hidden layer are plotted in Fig. 6 from the Table 1. It is required to fix up the number of neurons in the hidden layer of MLP so that it can show the optimal recognition performance on the test set.

Recognition performances of the MLP, as observed from the curve shown in Fig 6., initially rise as the number of neurons in the hidden layer is increased and falls after the same crosses some limiting value. It reflects the fact that for some fixed training and test sets, learning and generalization abilities of the MLP improve as the number of neurons in its hidden layer as increases up to certain limiting value and any further increase in the number of neurons in the hidden layer thereafter degrades the abilities. It is called the *over-fitting* problem.

The optimal recognition performance of the MLP is observed at a point, on the curve of Fig.6, where the number of neurons in its hidden layer is set to 60. Thus the number of neurons in the hidden layer of the MLP is finally fixed up to 60. With this, the design process is completed producing an MLP (76-60-50) for recognition of handwritten alphabet on the basis of the feature set explained before. Recognition performances of this MLP, as observed on the training and the test sets, are 86.46% and 75.05% respectively. Some samples of misclassified character images are shown in Fig. 7(a-d).

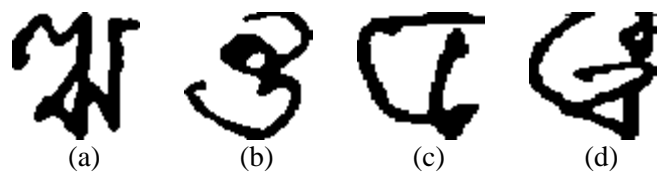

(a)        (b)        (c)        (d)

Fig. 7.  Some samples of misclassified character images.
(a) A character image of  misclassified as
(b) A character image of  misclassified as
(c) A character image of  misclassified as
(d) A character image of 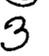 misclassified as

Recognition performances of the MLP can be further improved *firstly* by adding newer variation of handwritten alphabetic samples to the training set and *secondly* considering more discriminating features for characters. The work presented here can have useful application in the development of a complete OCR system for handwritten Bangla text.

## 5. ACKNOWLEDGEMENT